\newcommand{\xba}{\alpha}
\newcommand{\xbb}{\beta}

\newcommand{\xbd}{\delta}
\newcommand{\xbe}{\in}
\newcommand{\xbf}{\phi}
\newcommand{\xbg}{\gamma}

\newcommand{\xCK}{\times}

\newcommand{\xCN}{\neg}
\newcommand{\xCQ}{\emptyset}

\newcommand{\xCd}{\approx}

\newcommand{\xcA}{\forall}

\newcommand{\xcE}{\exists}

\newcommand{\xcK}{\not\leq}

\newcommand{\xcV}{\bigcup}

\newcommand{\xcc}{\subseteq}
\newcommand{\xcd}{\supseteq}
\newcommand{\xce}{\not\in}

\newcommand{\xcg}{\geq}
\newcommand{\xch}{\Rightarrow}

\newcommand{\xcj}{\Leftrightarrow}
\newcommand{\xck}{\leq}
\newcommand{\xcl}{\vdash}
\newcommand{\xcm}{\models}

\newcommand{\xco}{\vee}
\newcommand{\xcp}{\rightarrow}

\newcommand{\xcr}{\leftrightarrow}
\newcommand{\xcs}{\cap}
\newcommand{\xcu}{\wedge}
\newcommand{\xcv}{\cup}

\newcommand{\xcz}{\Box}

\newcommand{\xDd}{\equiv}

\newcommand{\xDl}{\hspace{0.5em}}

\newcommand{\xdl}{{\cal L}}

\newcommand{\xdp}{{\cal P}}

\newcommand{\xdy}{{\cal Y}}

\newcommand{\xEc}{\not<}
\newcommand{\xEd}{\neq}

\newcommand{\xec}{\preceq}

\newcommand{\xFO}{\parallel}

\newcommand{\xfA}{\mid}

\newcommand{\Xl}{\ldots}

\newcommand{\bl}{\begin{lemma} \rm}
\newcommand{\el}{\end{lemma}}
\newcommand{\br}{\begin{remark} \rm}
\newcommand{\er}{\end{remark}}
\newcommand{\be}{\begin{example} \rm}
\newcommand{\ee}{\end{example}}
\newcommand{\bco}{\begin{corollary} \rm}
\newcommand{\eco}{\end{corollary}}
\newcommand{\bc}{\begin{claim} \rm}
\newcommand{\ec}{\end{claim}}
\newcommand{\bfa}{\begin{fact} \rm}
\newcommand{\efa}{\end{fact}}
\newcommand{\bp}{\begin{proposition} \rm}
\newcommand{\ep}{\end{proposition}}
\newcommand{\bd}{\begin{definition} \rm}
\newcommand{\ed}{\end{definition}}
\newcommand{\bcs}{\begin{construction} \rm}
\newcommand{\ecs}{\end{construction}}
\newcommand{\bcd}{\begin{condition} \rm}
\newcommand{\ecd}{\end{condition}}
\newcommand{\bt}{\begin{theorem} \rm}
\newcommand{\et}{\end{theorem}}
\newcommand{\bn}{\begin{notation} \rm}
\newcommand{\en}{\end{notation}}
\newcommand{\bfi}{\begin{bild} \rm}
\newcommand{\efi}{\end{bild}}

\newcommand{\bfc}{\begin{figure} \begin{center}}
\newcommand{\efc}{\end{center} \end{figure}}

\documentstyle[12pt]{article}
\title{Distance Semantics for Belief Revision
\thanks{This work was partially supported
by the Jean and Helene Alfassa fund for
research in Artificial Intelligence and by grant 136/94-1 of the
Israel Science Foundation on ``New Perspectives on Nonmonotonic Reasoning''.}
}

\author{
Daniel Lehmann\thanks{Institute of Computer Science, Hebrew University,
91904 Jerusalem, Israel, lehmann@cs.huji.ac.il}
\and
Menachem Magidor\thanks{Institute of Mathematics, Hebrew University,
91904 Jerusalem, Israel, menachem@math.huji.ac.il}
\and
Karl Schlechta\thanks{
Laboratoire d'Informatique de Marseille, CNRS ESA 6077,
CMI, 39 rue Joliot Curie,
F-13453 Marseille C\'{e}dex 13, France, ks@gyptis.univ-mrs.fr}
}

\date{August 3, 1999}

\begin{document}

\newtheorem{lemma}{Lemma}[section]
\newtheorem{theorem}[lemma]{Theorem}
\newtheorem{proposition}[lemma]{Proposition}
\newtheorem{corollary}[lemma]{Corollary}
\newtheorem{claim}[lemma]{Claim}
\newtheorem{fact}[lemma]{Fact}
\newtheorem{remark}[lemma]{Remark}
\newtheorem{definition}{Definition}[section]
\newtheorem{construction}{Construction}[section]
\newtheorem{condition}{Condition}[section]
\newtheorem{example}{Example}[section]
\newtheorem{notation}{Notation}[section]
\newtheorem{bild}{Figure}[section]

\maketitle

\begin{abstract}

A vast and interesting family of natural semantics for belief revision
is defined.
Suppose one is given a distance $d$ between any two models.
One may then define the revision of a theory $K$ by a formula
$ \xba $ as the theory defined by the set of all those models of $ \xba $
that are closest, by $d$, to the set of models of $K$.
This family is characterized by a set of rationality postulates that
extends the AGM postulates.
The new postulates describe properties of iterated revisions.

\end{abstract}

\unitlength1.0mm

\newsavebox{\EINSeins}
\savebox{\EINSeins}(130,50)[bl]
{

\put(29,45){Figure 1.1}

\put(10,20){\circle*{1.5}}
\put(30,20){\circle*{1.5}}
\put(70,20){\circle*{1.5}}
\put(70,20){\circle{20}}

\put(10,20){\line(1,0){70}}

\put(10,30){-3}
\put(30,30){-2}
\put(70,30){0}

\put(10,08){$m_1$}
\put(30,08){$m_2$}
\put(70,08){$m$}
\put(77,13){$X$}

}

\newsavebox{\ZWEIeins}
\savebox{\ZWEIeins}(130,85)[bl]
{

\put(29,80){Figure 2.1}

\put(5,50){\line(1,0){30}}
\put(35,50){\line(1,2){6}}
\put(35,50){\line(1,-2){6}}

\put(5,50){\circle*{1.5}}
\put(35,50){\circle*{1.5}}

\put(41,62){\circle*{1.5}}
\put(41,38){\circle*{1.5}}

\put(5,47){$x$}
\put(32,47){$y$}
\put(43,61){$a$}
\put(43,37){$b$}

\put(65,50){\line(1,0){35}}
\put(100,50){\line(1,2){12}}
\put(100,50){\line(1,-2){12}}

\put(65,50){\circle*{1.5}}
\put(100,50){\circle*{1.5}}

\put(112,74){\circle*{1.5}}
\put(112,26){\circle*{1.5}}

\put(65,47){$x$}
\put(97,47){$y$}
\put(114,73){$a'$}
\put(114,25){$b'$}

}

\newsavebox{\ZWEIzwei}
\savebox{\ZWEIzwei}(130,72)[bl]
{

\put(29,65){Figure 2.2}

\put(5,35){Case 2:}

\put(5,20){\line(1,0){100}}

\put(5,20){\vector(1,0){11}}
\put(17,20){\vector(1,0){7}}
\put(25,30){\vector(-1,0){20}}
\put(5,30){\vector(0,-1){9}}
\put(25,20){\vector(0,1){10}}

\put(55,20){\vector(1,0){10}}
\put(55,20){\vector(-1,0){10}}

\put(95,20){\vector(1,0){7}}
\put(95,20){\vector(-1,0){7}}

\put(5,20){\circle*{1.5}}
\put(17,20){\circle*{1.5}}
\put(25,20){\circle*{1.5}}
\put(85,20){\circle*{1.5}}
\put(105,20){\circle*{1.5}}

\put(5,15){0}
\put(17,15){1}
\put(25,15){1.1}
\put(85,15){16}
\put(105,15){20}

\put(5,10){$a'$}
\put(17,10){$a$}
\put(85,10){$b$}
\put(105,10){$b'$}

\put(5,55){Case 1:}

\put(5,50){\line(1,0){100}}

\put(5,50){\circle*{1.5}}
\put(17,50){\circle*{1.5}}
\put(85,50){\circle*{1.5}}
\put(105,50){\circle*{1.5}}

\put(5,45){0}
\put(17,45){1}
\put(85,45){16}
\put(105,45){20}

\put(5,40){$a$}
\put(17,40){$a'$}
\put(85,40){$b$}
\put(105,40){$b'$}

}

\section{
Introduction
}


\subsection{
Overview and related work
}


The aim of this paper is to investigate semantics and
logical properties of theory revisions based on an underlying notion
of distance between individual models.
In many situations it is indeed reasonable to assume that the agent
has some natural way to evaluate the distance between any two models
of the logical language of interest.
The distance between model $m$ and model $m'$ is a measure of how
far $m'$ appears to be
from the point of view of $m$.
This distance may measure how different $m'$ is from $m$ under some objective
measure: e.g., the number of propositional atoms on which $m'$ differs from
$m$ if the language is propositional and finite,
but it may also reflect a subjective assessment of the agent about
its own capabilities such as, for example, the probability that,
if $m$ is the case,
the agent (wrongly) believes that $m'$ is the case.
Any such distance between models may be used to define a procedure
for theory revision: both a theory $T$ and a formula $\alpha$ define
a set of models, $M(T)$ and $M(\alpha)$, respectively,
and the result of revising $T$ by $\alpha$, $T*\alpha$, is the theory
of the set of those models of $M(\alpha)$ that are closest to $M(T)$.

The purpose of this paper is not to suggest specific useful
notions of distance.
It assumes some abstract notion of a
distance is given and studies the properties
of the revisions defined by this distance.
The distances that will be considered in this paper do not always
satisfy the properties generally accepted for distances, they are
really pseudo-distances.
There is no obvious reason why, in particular, our distance should
satisfy the triangular inequality or even be symmetric.
It may be the case that, from the perspective of $m$, $m'$ looks very
far away, but, from the perspective of $m'$, $m$ looks close by.
In the terms of one of our examples above: if $m$ is the case,
our agent may give a very low probability to $m'$ being the case,
but if $m'$ is the case, our agent may well be hesitant about whether
$m$ or $m'$ is the case: assume, for example, that $m$ and $m'$
differ by the value of one atomic proposition $p$ that is tested
by our agent. The test for $p$ may well be very reliable if $p$ is the case
but quite unreliable if $p$ is not the case.

\subsubsection{
Related work
}
In [AGM85], Alchourr\'{o}n, G\"{a}rdenfors and Makinson introduced the study
of theory revision.
Their account of revision is indirect: they describe contractions in terms
of maximal non-implying sub-theories and they go on to characterize revisions,
reducing revision to contraction via the Levi identity.
In [Gro88], Grove gave a direct, semantic, characterization of revision.
The result of revising a theory $K$ by a proposition $\alpha$
is determined by the models of $\alpha$ that are, individually,
{\em closest} to the set, taken collectively, of all models of $K$.
It thus uses a relation between individual models and sets of
models.
It is natural to seek to analyze such {\em closeness} in terms of
a distance function between models.
A first attempt was made by Becher in [Bec95],
in view of comparing revision and update in a unified setting.
Becher worked with not necessarily symmetric distances
and showed that the AGM postulates hold in distance
based revision, but gave no representation result.
Independently, the authors presented, in [SLM96],
a preliminary version of the results given in this paper.
There, only the finite case of symmetric distances was treated.
We deal here with the infinite case of symmetric distances and
with the finite case of non-symmetric distances.
We also provide here proofs and counter-examples.
We present, first, our results on an abstract level,
dealing with abstract sets and, then, specialize our results
to the case of sets of models.
In recent work (personal communication) Areces and Becher
gave a representation result for the arbitrary, i.e. infinite and
not necessarily symmetric case. Their conditions are different from
ours, based on complete consistent theories, i.e.
single models, and partly in an ``existential'' style, whereas
our conditions are ``universal'' and more in the AGM style.
We do not know whether there is an easy direct, i.e. not going via the
semantics, proof of the equivalence of the Areces/Becher and our
conditions. Thus, their approach is an alternative route, whose
relation to our results is a subject of further research.
The infinite case for non-symmetric distances with conditions
in our style is still open.

\subsubsection{
Structure of this paper
}

After a short motivation in Section 1.1, we present and discuss the AGM
framework for revision in Section 1.2 and modify it slightly.
Section 1.3 introduces pseudo-distances, which are distances weakened to
the
properties essential in our context. In particular, pseudo-distances are
not
necessarily symmetric.
We formalize
revision based on a (pseudo-) distance, and we show that the usual AGM
properties
hold for such distance-based revisions, at least in the finite case. An
additional
property (definability preservation) guarantees them to hold in the
infinite
case, too. Here, we also discuss some
properties of distance-based revision going beyond the AGM postulates.

Our main results are algebraic in nature, and work for arbitrary sets, not
only for sets of models. The translation to logic is then straightforward.

Section 2 presents the algebraic representation results, which describe
the
conditions which guarantee that a binary set operator $ \xfA $ is
representable by a
pseudo-distance $d,$ i.e. that $A \xfA B$ is the set of $b \xbe B$
$d$-closest to $A$, formally that
$A \xfA B$ $=$ $A \xfA_{d}B$ $:=$ $\{b \xbe B:$ $ \xcE a_{b} \xbe A
\hspace{0.2em} \xcA
a' \xbe A \hspace{0.2em} \xcA b' \xbe B.d(a_{b},b) \xck d(a',b' )\}.$

In Section 2.2, we treat the case of symmetric pseudo-distances, the
result applies to sets of arbitrary cardinality.
Note that this infinite case requires a limit condition: For non-empty sets $A$,
$B$ there is some $b \xbe B$ with closest distance (among the elements of $B$)
to $A$.
In Section 2.3, we treat the not
necessarily symmetric case, but our result applies only to the finite
situation.

Section 3 finally translates the results of Section 2 to logic. We there
describe
the conditions which guarantee that a revision operator $*$ can be
represented
by a pseudo-distance $d$ between models, i.e. $T*T':=Th(M(T)
\xfA_{d}M(T' ))$ - where
$M(T)$ is the set of models of the theory $T,$ and $Th(X)$ the set of
formulas
valid in the set of models $X.$

Analogously to the algebraic characterization, the logical representation
results are for possibly infinite languages in the symmetric case (with
some
caveat about definability preservation) and for finite ones in the not
necessarily symmetric case.

\subsection{
Belief revision
}

Intelligent agents must gather information about the world,
elaborate theories about it and revise those theories in view of new
information that, sometimes, contradicts the beliefs previously held.
Belief revision is therefore a central topic in Knowledge Representation.
It has been studied in different forms: numeric or symbolic, procedural or
declarative, logical or probabilistic.

\subsubsection{
The AGM framework
}


One of the most successful frameworks in which belief revision has been
studied
has been proposed by Alchourr\'{o}n, G\"{a}rdenfors and Makinson, and is
known as the AGM framework. It deals with operations of revision that
revise a theory (the set of previous beliefs) by a formula
(the new information). It proposes a set of rationality postulates that
any
reasonable revision should satisfy.
A large number of researchers in AI have been attracted by and have
developed this approach further: both in the abstract and by devising
revision procedures that satisfy the AGM rationality postulates.

Two remarks should be made immediately.
First, the AGM framework presents rationality postulates for revision. It does
not choose anyone specific revision among the many possible revisions
that satisfy those postulates.
Those postulates are justified and defended by the authors, but, recently,
some doubts have been expressed as to their desirability, at least for
modeling updates, see [KM92] and, more importantly for us,
it is not clear that the AGM postulates are all what one
would like. A number of authors,
in particular [FL94], [DP94], [Leh95], have in fact
argued that one would expect some additional postulates to hold.
But the consideration of additional postulates has proved slippery
and dangerous: the postulates proposed in [DP94] have been shown
inconsistent in [Leh95]
and have been modified in [DP97]. But this modification forces on us
the rejection of one of the basic ontological commitments of the AGM framework,
which brings us to a second remark.
Secondly, one of the basic ontological commitments of the AGM approach
is that what the agent is revising is a belief set. In other terms,
epistemic states are belief sets.
That this is the AGM
position is clear from the formalism chosen:
the left hand argument of the star revision operation is a belief set,
from the motivation presented, and it is explicitly recognized
in [G"r88], p. 47.

Over the years, a large number of researchers have moved away from this
identification, sometimes without recognizing
it [BG93], [Bou93], [DP97], [Wil94], [NFPS95].
Recently, conclusive evidence has been put
forward [Leh95], [FH96] to the effect that this identification
of epistemic states
to belief sets is not welcome in many AI applications.
When iterated revisions are considered,
it is reasonable to assume that the agent's epistemic state includes
information related to its history of revisions and that all this history,
not only the agent's current belief set, may influence future revisions.

This paper keeps the AGM commitment to identifying epistemic states with
belief sets, but proposes additional rationality postulates. Those additional
postulates characterize exactly the revisions that are defined
by pseudo-distances. They constrain revisions in the way they treat
their left argument, the theory to be revised (in this respect the
AGM postulates are extremely, probably excessively, liberal)
and they imply highly non-trivial
properties for iterated revisions.
This paper therefore treats iterated revisions within the original
AGM commitment to the identification of epistemic states and belief sets.
Results related to the ideas of this paper may be found in [BGHPSW97].
Similar ideas in a context in which epistemic states are not belief sets
may be found in [BLS99].

This work provides a semantics for theory revision \`a la AGM, or for a
sub-family of such revisions.
It is the first such effort to describe semantically the whole
revision operation $*$ in a unified way.
Previous attempts [Gro88], [GM88] describe the revision
of each theory $K$ by a different structure without any {\em glue}
relating the different structures: sphere systems or epistemic entrenchment
relations, corresponding to different $K$'s.
In this paper, the revisions of the different $K$'s are obtained from the same
pseudo-distance. A tight fit (coherence) between the revisions
of different $K$'s seem crucial for a useful treatment of iterated
revisions: it must be the same revision operation that executes the
successive revisions for any interesting properties to appear.
Our revisions are therefore defined by a polynomial (in the number
of models considered) number of pseudo-distances instead of an exponential
number of sphere systems or epistemic entrenchment relations (one for
each theory $K$).

Semantics based on a more or less abstract notion of distance
is not a new idea in non-classical logics.
The best known example is perhaps the Stalnaker/Lewis
distance semantics for counterfactual conditionals, see e.g. [Lew73].

The AGM framework, defined in [AGM85],
studies revision operations, denoted $*,$ that operate
on two arguments: a set $K$ of formulas closed under logical deduction
on the left and a formula $ \xba $ on the right. Thus $K* \xba $ is the
result of revising
theory $K$ by formula $ \xba,$ using revision method $*.$

\bn

$\hspace{0.5em}$


Our base logic will be classical propositional logic, though our main
results
are purely algebraic in nature, and therefore carry over to other logics,
too.

By abuse of language, we call a language finite whose set of propositional
variables is finite.

A theory will be an arbitrary set of formulas, not necessarily deductively
closed.

We use the customary notation $Cn(T)$ for the set of all logical
consequences of
a theory $T.$ $Cn(T, \xba )$ will stand for $Cn(T \xcv \{ \xba \}).$

$Con(T)$ will stand for: $T$ is (classically) consistent, $Con(T,T' )$
abbreviates
$Con(T \xcv T' ).$

$ \xcm $ will be classical validity, and $ \xcm T \xcr T' $ will
abbreviate the obvious:
$ \xcm T \xcp \xbf ' $ for all $ \xbf ' \xbe T' $ and $ \xcm T'
\xcp \xbf $ for all $ \xbf \xbe T.$

Given a propositional language $ \xdl,$ $M_{ \xdl }$ will be the set of
its models.

$M(T)$ will be the models of a theory $T$ (likewise $M( \xbf )$ for a
formula $ \xbf ),$ and
$Th(X)$ the set of formulas valid in a set of models $X.$

$T \xco T' $ $:=$ $\{ \xbf \xco \xbf ':$ $ \xbf \xbe T,$ $ \xbf '
\xbe T' \}.$

$ \xdp $ will be the power set operator.

The logical connectives $ \xcu $ and $ \xco $ and the set connectives $
\xcs $ and $ \xcv $ always have
precedence over the revision and set operators $*$ and $ \xfA.$

Numbering of conditions:
$( \xfA i)$ will number conditions common to the symmetric and the not
necessarily
symmetric set operators $ \xfA,$ $( \xfA Si)$ and $( \xfA Ai)$ conditions
for respectively the
symmetric and the not necessarily symmetric set operator $ \xfA.$
$(*i),$ $(*Si),$ $(*Ai)$ will do the same for the theory revision operator
$*.$

\en

The original AGM rationality postulates are the following, for $K$ a
deductively
closed set of formulas, and $ \xba,$ $ \xbb $ formulas.

\bd

$\hspace{0.5em}$


$(K*1)$ $K* \xba $ is a deductively closed set of formulas.

$(K*2)$ $ \xba \xbe K* \xba.$

$(K*3)$ $K* \xba \xcc Cn(K, \xba ).$

$(K*4)$ If $ \xCN \xba \xce K,$ then $Cn(K, \xba ) \xcc K* \xba.$

$(K*5)$ If $K* \xba $ is inconsistent then $ \xba $ is a logical
contradiction.

$(K*6)$ If $ \xcm \xba \xcr \xbb $, then $K* \xba =K* \xbb.$

$(K*7)$ $K* \xba \xcu \xbb \xcc Cn(K* \xba, \xbb ).$

$(K*8)$ If $ \xCN \xbb \xce K* \xba,$ then $Cn(K* \xba, \xbb ) \xcc K*
\xba \xcu \xbb.$

\ed

\subsubsection{
Modifications of the AGM framework
}


We prefer to modify slightly the original AGM formalism on two accounts.
First, it seems to us that the difference required in the types of the
two arguments of a revision: the left argument being a theory
and the right argument being a formula is not founded.
The lack of symmetry is twofold: the left-hand argument, being a theory,
may be inherently infinite and not representable by a single formula
while the right-hand argument is always a single formula,
but also the left-hand argument is not an arbitrary set of formulas, but
closed under logical implication, whereas the right-hand argument is not
deductively closed, thus requiring Postulate
$K*6$ to assert invariance under logical equivalence.
There is no serious reason for this lack of symmetry.
We shall therefore prefer a formalism that is symmetric in both arguments.

Our results for symmetric pseudo-distances are valid for infinite sets,
our
results for not necessarily symmetric pseudo-distances have been proved
only
for finite sets - where sets are to
be understood as sets of models. The latter are thus proved for languages
based
on a finite set of propositional variables only. We therefore choose a
formalism
in which both arguments of the revision operator are theories, which, in
the
not necessarily symmetric case, will be assumed to be
equivalent to single formulas.
We thus look at $T*T',$ the theory that is the result of revising
theory $T$ by the new information represented by theory $T'.$

Secondly, in the AGM formalism, each one of the theory $K$
and the formula $ \xba $ may be inconsistent.
There is no harm in doing so, but the interesting revisions
are always revisions of consistent theories by consistent formulas
and the consideration of inconsistent arguments makes the treatment
unnecessarily clumsy.
Therefore, we shall only revise consistent theories by consistent
theories,
and assume both arguments are consistent.

The AGM postulates may now be rewritten in the following way.
We rewrite $K*3$ and $K*4$ in one single postulate, $(*3),$
and similarly for $K*7$ and $K*8,$ in $(*4).$ $K*1$ and $K*5$ are
summarized in our
general prerequisite and $(*1).$

Remember that $T,T',T'',S,S'$ are now arbitrary consistent theories.

\bd

$\hspace{0.5em}$


$(*0)$ If $ \xcm T \xcr S,$ $ \xcm T' \xcr S',$ then $T*T' =S*S'
,$

$(*1)$ $T*T' $ is a consistent, deductively closed theory,

$(*2)$ $T' \xcc T*T',$

$(*3)$ If $T \xcv T' $ is consistent, then $T*T' =Cn(T \xcv T' ),$

$(*4)$ If $T*T' $ is consistent with $T'',$ then $T*(T' \xcv
T'' )=Cn((T*T' ) \xcv T'' ).$

\ed

\subsection{
Revision based on pseudo-distances
}

\subsubsection{
Pseudo-distances
}

We will base our semantics for revision on pseudo-distances between
models.
Pseudo-distances differ from distances in that their
values are not necessarily reals, no addition of values has to be defined,
and symmetry need not hold. All we need is a totally ordered set of
values.
If there is a minimal element 0 such that $d(x,y)=0$ iff $x=y,$ we say
that $d$
respects identity. Pseudo-distances which do not respect identity have
their
interest in situations where staying the same requires effort.

We first recollect:

\bd

$\hspace{0.5em}$


A binary relation $ \xck $ on $X$ is a preorder, iff $ \xck $ is
reflexive and transitive. If $ \xck $ is in
addition total, i.e. iff $ \xcA x,y \xbe X$ $x \xck y$ or $y \xck x,$ then
$ \xck $ is a total preorder.

A binary relation $<$ on $X$ is a total order, iff $<$ is transitive,
irreflexive,
i.e. $x \xEc x$ for all $x \xbe X,$ and for all $x,y \xbe X$ $x<y$ or
$y<x$ or  $x=y.$

\ed

\paragraph{
Note 1.1:
}

$\hspace{0.5em}$

If $ \xck $ is a total preorder on $X,$ $ \xCd $ the corresponding
equivalence relation
defined by $x \xCd y$ iff $x \xck y$ and $y \xck x,$ [x] the $ \xCd $
-equivalence
class of $x,$ and we define $[x]<[y]$ iff $x \xck y,$ but not $y \xck x,$
then $<$ is a total
order on $\{[x]:x \xbe X\}.$

\bd

$\hspace{0.5em}$


$d:U \xCK U \xcp Z$ is called a pseudo-distance on $U$ iff (d1) holds:

(d1) $Z$ is totally ordered by a relation $<.$

If, in addition, $Z$ has a $<$ -smallest element 0, and (d2) holds, we say
that $d$
respects identity:

(d2) $d(a,b)=0$ iff $a=b.$

If, in addition, (d3) holds, then $d$ is called symmetric:

(d3) $d(a,b)=d(b,a).$

(For any $a,b \xbe U.)$

Let $ \xck $ stand for \mbox{$< \cup =$}.

\ed

Note that we can force the triangle inequality to hold trivially (if we
can
choose the values in the real numbers): It suffices to choose the values
in
the set $\{0\} \xcv [0.5,1],$ i.e. in the interval from 0.5 to 1, or as 0.

Recall that our main representation results are purely algebraic, and
apply to
arbitrary sets $U,$ which need not necessarily be sets of models.
Intuitively however, $U$ is to be understood as the set of models for some
language $ \xdl,$ and the distance from $m$ to $n,$ $d(m,n)$ represents
the ``cost'' or the ``difficulty'' of a change from the situation
represented
by $m$ to the situation represented by $n.$
M. Dalal [Dal88] has considered one such distance: the distance between
two propositional worlds is the number of atomic propositions on which
they differ, i.e., the Hamming distance between worlds considered as
binary k-dimensional vectors, where $k$ is the number of atomic
propositional variables.
A. Borgida [Bor85] considered a similar but different distance,
based on set inclusion. His distances are not totally ordered and
therefore the
framework presented here does not fit his work.

Another example of such a distance is the trivial distance:
$d(m,n)$ is 0 if $m=n$ and 1 otherwise.

Both those distances satisfy the triangular inequality.
In applications dealing with reasoning about actions and change,
one may want to consider the distance between two models to represent
how difficult, or unexpected, the transition is.
In such a case, a natural pseudo-distance may well not be symmetric.

We give the formal definition of the elements of $B$ d-closest to $A$:

\bd

$\hspace{0.5em}$


Given a pseudo-distance $d:U \xCK U \xcp Z$, let for $A,B \xcc U$
$A \xfA_{d}B$ $:=$ $\{b \xbe B:$ $ \xcE a_{b} \xbe A \xcA a' \xbe A \xcA
b' \xbe B.d(a_{b},b) \xck d(a',b' )\}$.

\ed

Definition 1.5 may be presented in a slightly different light.
Put \mbox{$(a , b) < (a' , b')$} iff \mbox{$d(a , b) < d(a' , b')$}.
Let \mbox{$min_{<}(A \times B)$} be the set of all minimal elements
(under $<$) of the set \mbox{$A \times B$}. Then, $A \xfA_{d}B$ is nothing else
than the right projection (on $B$) of \mbox{$A \times B$}.

Thus, $A \xfA_{d}B$ is the subset of $B$ consisting of all $b \xbe B$ that
are closest to $A$.
Note that, if $A$ or $B$ is infinite, $A \xfA_{d}B$ may be empty, even if $A$
and $B$ are not
empty. A condition assuring non-emptiness will be imposed when necessary.

The aim of Section 2 of this article is to characterize those operators
$ \xfA: \xdp (U) \xCK \xdp (U) \xcp \xdp (U),$ for which there is a
pseudo-distance $d,$ such that
$A \xfA B=A \xfA_{d}B.$ We call such $ \xfA $ representable:

\bd

$\hspace{0.5em}$


An operation $ \xfA $ is representable iff there is a pseudo-distance
$d:U \xCK U \xcp Z$ such that

(1) $A \xfA B$ $=$ $A \xfA_{d}B$ $:=$ $\{b \xbe B:$ $ \xcE a_{b} \xbe A
\xcA a' \xbe A \xcA b' \xbe B(d(a_{b},b) \xck d(a',b' ))\}.$

\ed

\subsubsection{
Revision based on pseudo-distances
}


The representation results of [AGM95],
the semantics of Grove [Gro88] and
the very close connection with the rational relations of [LM92],
showed in [GM94], all leave essentially unanswered the question
of the nature of the dependence of the
revision $T*T'$ on its left argument, $T$.
Since we, like most researchers in Artificial Intelligence, are mostly
interested in iterated revisions, proper understanding, and semantics,
for this dependence is crucial.
The purpose of this paper is to answer the question by
proposing a suitable semantics.
We completely characterize the semantics by a set of postulates.
We do not claim that the semantics proposed are the most general ones,
we present one family of reasonable semantics, based on pseudo-distances
between models.

The following is the central definition, it describes the way a revision
$*_{d}$ is
attached to a pseudo-distance $d$ on the set of models.

\bd

$\hspace{0.5em}$


$T*_{d}T' $ $=$ $Th(M(T) \xfA_{d}M(T' )).$

$*$ is called representable iff there is a pseudo-distance $d$ on the set
of models
s.t. $T*T' =Th(M(T) \xfA_{d}M(T' )).$

\ed

The main goal of this work is to characterize the properties, i.e.,
rationality
postulates satisfied by revisions representable by pseudo-distances.

\subsubsection{
Revision based on pseudo-distances and the AGM postulates
}

\paragraph{
The AGM postulates hold for revision based on pseudo-distances in the
finite case
}

$\hspace{0.5em}$

\bd

$\hspace{0.5em}$


An operation $ \xfA $ on the sets of models of some logic is called
definability
preserving iff $M(T) \xfA M(T' )$ is again the set of models of some
theory $S$ for all
theories $T,$ $T'.$

\ed

Abstractly, definability preservation strongly couples proof theory and
semantics. To obtain the same kind of results without definability
preservation,
we would have to allow a ``decoupling'' on a ``small'' set of exceptions. This
is
illustrated e.g. by the results in [Sch97] for the definability
preservation
case, and in [Sch98] for the unrestricted case of representation
results for
preferential structures. A similar problem arose already in a finite
situation
in [ALS98] in the context of partial and total orders, and is treated
there
by an inductive process.

A first easy result is: any such revision defined for a $finite$ language
satisfies the AGM postulates $(*0)-(*4),$ if $d$ respects identity.
(We use $ \xfA $ to abbreviate $ \xfA_{d}.)$

The same proof shows that the AGM postulates also hold in the infinite case,
if the operation $\xfA$ is definability preserving, and if we impose a
limit condition for postulate $(*1)$.

$(*0)$ is evident, as we work with models.

$(*1)$ holds in the finite case, we will impose it, i.e. a limit
condition, in the infinite case.

$(*2)$ trivial by definition.

$(*3)$ this holds, as $d(a,a)$ is minimal for all $a$, by respect of
identity.

$(*4)$ Note that $M(S \xcv S' )=M(S) \xcs M(S' ),$ and that $M(S*S'
)=M(S) \xfA M(S' ).$
By prerequisite, $M(T*T' ) \xcs M(T'' ) \xEd \xCQ,$
so $(M(T) \xfA M(T' )) \xcs M(T'' ) \xEd \xCQ.$
Let $A:=M(T),$ $B:=M(T' ),$ $C:=M(T'' ).$ `` $\xcc $'': Let
$b \xbe A \xfA (B \xcs C).$
By prerequisite, there is $b' \xbe (A \xfA B) \xcs C.$ Thus $d(A,b' )
\xcg d(A,B \xcs C)=d(A,b).$
As $b \xbe B,$ $b \xbe A \xfA B,$ but $b \xbe C,$ too. `` $\xcd $''
: Let $b' \xbe (A \xfA B) \xcs C.$ Thus $d(A,b' )=$
$d(A,B) \xck d(A,B \xcs C),$ so by $b' \xbe B \xcs C$ $b' \xbe A \xfA
(B \xcs C).$
We conclude $M(T) \xfA (M(T' ) \xcs M(T'' ))$ $=$ $(M(T) \xfA M(T'
)) \xcs M(T'' ),$ thus that
$T*(T' \xcv T'' )=Cn((T*T' ) \xcv T'' ).$

\paragraph{
The AGM postulate $(*4)$ may fail in the infinite not
definability preserving case
}

$\hspace{0.5em}$

The importance of definability preservation is illustrated by the
following
example, which shows that already the AGM properties may fail when the
distance
between models does not preserve definability. Essentially the same
example
will show in Section 3 (Example 3.1 there) that our Loop Condition $(*S1)$
may
fail when the distance is not definability preserving. We see here that
this
is not related to our stronger conditions, but happens already in the
general
AGM framework.

\be

$\hspace{0.5em}$


Consider an infinite propositional language $ \xdl.$

Let $T,$ $T_{1},$ $T_{2}$ be complete (consistent) theories, $T' $ a
theory with infinitely
many models, $M(T)=\{m\},$ $M(T_{1})=\{m_{1}\},$ $M(T_{2})=\{m_{2}\},$
$M(T' )=X \xcv \{m_{1},m_{2}\},$ $M(T'' )=\{m_{1},m_{2}\}.$
Assume further $Th(X)=T',$ so $X$ is not definable by a theory.

Arrange the models of $ \xdl $ in the real plane s.t. all $x \xbe X$ have
the same
distance $<2$ (in the real plane) from $m,$ $m_{2}$ has distance 2 from $m,$
and $m_{1}$ has
distance 3 from $m.$

(See Figure 1.1.)

Then $M(T) \xfA M(T' )=X,$ but $T*T' =T',$ so $T*T' $ is
consistent with $T'',$ and
$Cn((T*T' ) \xcv T'' )=T''.$ But $T' \xcv T'' =T'',$
and $T*(T' \xcv T'' )=T_{2} \xEd T''.$
$ \xcz $
\\[3ex]

\ee


\bfc
\begin{picture}(110,50)
\put(0,0){\usebox{\EINSeins}}
\end{picture}
\efc

\paragraph{
AGM revisions are not all definable by pseudo-distances
}

$\hspace{0.5em}$

But any revision defined by a pseudo-distance $d$ also satisfies
some properties that do not follow from the AGM postulates. We note again
$ \xfA $ for $ \xfA_{d}.$

Consider, for example, the set $C=(A_{1} \xcv A_{2}) \xfA B,$ where
$A_{i}$ and $B$ are finite sets.
$d(A,B)$ will be $min\{d(a,b):a \xbe A,b \xbe B\}.$

If $d(A_{1},B)<d(A_{2},B),$ then $C=A_{1} \xfA B.$
If $d(A_{2},B)<d(A_{1},B),$ we have $C=A_{2} \xfA B.$
If $d(A_{1},B)=d(A_{2},B),$ then we have $C=(A_{1} \xfA B) \xcv (A_{2}
\xfA B).$
It follows that any revision defined by a pseudo-distance satisfies (for a
finite language):
$( \xba_{1} \xco \xba_{2})* \xbb $ is equal to $( \xba_{1}* \xbb ) \xcs (
\xba_{2}* \xbb ),$ to $ \xba_{1}* \xbb,$ or to $ \xba_{2}* \xbb.$

This property does not follow from the AGM postulates, as will be shown
below, but seems a very natural property.
Indeed, when revising a disjunction
$ \xba_{1} \xco \xba_{2}$ by a formula $ \xbb,$ there are two
possibilities.
First, it may be the case that our indecision concerning
$ \xba_{1}$ or $ \xba_{2}$ persists after the revision, and, in this case,
the
revised theory is naturally the disjunction of the revisions.
But it may also be the case that the new
information $ \xbb $ makes us revise backwards and conclude that it must
be the case that $ \xba_{1}$ or, respectively, $ \xba_{2}$ was
(before the new information) the better theory and, in this case,
the revised theory should be $ \xba_{1}* \xbb $ or $ \xba_{2}* \xbb.$
Notice that this last property of revisions generated by
pseudo-distances is the left argument analogue
of AGM's Ventilation Principle which concerns the argument
on the right. The Ventilation Principle follows from the AGM postulates
and states that:
$ \xba *( \xbb_{1} \xco \xbb_{2})$
is equal to $( \xba * \xbb_{1}) \xcs ( \xba * \xbb_{2})$,
to $ \xba * \xbb_{1}$ or to $ \xba * \xbb_{2}$.

One can conclude that any revision defined by a pseudo-distance satisfies
the following properties, that deal with iterated revisions:

if $ \xbd \xbe (K* \xba )* \xbg $ and $ \xbd \xbe (K* \xbb )* \xbg,$ then
$ \xbd \xbe (K*( \xba \xco \xbb ))* \xbg $

and

if $ \xbd \xbe (K*( \xba \xco \xbb ))* \xbg,$ then, either $ \xbd \xbe
(K* \xba )* \xbg $ or $ \xbd \xbe (K* \xbb )* \xbg.$

Those properties seem intuitively right.
If after any one of two sequences of revisions that differ only at step
$i$
(step $i$ being $ \xba $ in one case and $ \xbb $ in the other),
one would conclude that $ \xbd $ holds, then one should
conclude $ \xbd $ after the sequence of revisions that differ from
the two revisions only in that step $i$ is a revision by
the disjunction $ \xba \xco \xbb,$ since knowing which of $ \xba $ or $
\xbb $ is true cannot be crucial.
This property is an analogue for the left argument of the Or property
of [KLM90].
Similarly, if one concludes $ \xbd $ from a revision by a disjunction,
one should conclude it from at least one of the disjuncts.
This property is an analogue for the left argument of the
Disjunctive Rationality property
of [KLM90], studied in [Fre93].
It is easy to see that the property (C1) of
Darwiche and Pearl [DP94], i.e., $(K* \xba )*( \xba \xcu \xbb )$ $=$ $K*(
\xba \xcu \xbb )$
is not satisfied by all revisions defined by pseudo-distances.
Section 2 will precisely characterize those revisions that
are defined by pseudo-distances.

Notice that in each of the AGM postulates, the left-hand side argument
of the revision operation $(*)$ is the same all along: all revisions
have the form $K*.$
Since, as has been shown above, every revision defined by a
pseudo-distance
satisfies the AGM postulates, if, for each theory $K$ we define
$K*$ by some pseudo-distance, then the revision defined
will satisfy the AGM postulates, even if we use different pseudo-distances
for different $K$'s.

Consider, for a simple example, 4 points in the real plane, $a$, $b,$ $c,$
$d,$ to
be interpreted as the models of a propositional language of two variables.
Let $a$ have the coordinates (0,1), $b$ (0,-1), $c$ (1,0), $d$ (2,0), and
define
by the natural distance the revisions with any $X \xEd \xCQ $ except for
$X:=\{a,b\}$
on the left hand side. As seen above, they will satisfy the AGM
postulates.
To define the revisions with $\{a,b\}$ on the left hand side, interchange
the
positions of $c$ and $d.$ This, too, satisfies the AGM postulates. As the
AGM
postulates say nothing about coherence between different $K$'s, all these
revisions together satisfy the AGM postulates.

But we will then have $\{a\} \xfA \{c,d\}=\{b\} \xfA \{c,d\}=\{c\},$ but
$\{a,b\} \xfA \{c,d\}=\{d\}$,
so such a system of revisions cannot be defined by a pseudo-distance.

\section{
The algebraic representation results
}


\subsection{
Introduction
}


First, a generalized abstract nonsense result.
This result is certainly well-known and we claim no priority.
It will be used repeatedly below to extend a relation $R$
to
a relation $S$.
The equivalence classes under $S$ will be used to define the
abstract distances.

\bl

$\hspace{0.5em}$


Given a set $X$ and a binary relation $R$ on $X,$ there exists a total
preorder $S$ on
$X$ that extends $R$ such that

(2) $ \xcA x,y \xbe X(xSy,ySx \xch xR^{*}y)$

where $R^{*}$ is the reflexive and transitive closure of $R.$

\el

\paragraph{
Proof:
}

$\hspace{0.5em}$

Define $x \xDd y$ iff $xR^{*}y$ and $yR^{*}x$.
The relation $ \xDd $ is an equivalence relation.
Let $[x]$ be the equivalence class of $x$ under $ \xDd$.
Define $[x] \xec [y]$ iff $xR^{*}y$.
The definition of $ \xec $ does not depend on the representatives
$x$ and $y$ chosen.
The relation $ \xec $ on equivalence classes is a partial order:
reflexive, antisymmetric and transitive.
A partial order may always be extended to a total order.
Let $ \xck $ be any total order on these equivalence classes
that extends $ \xec$.
Define $xSy$ iff $[x] \xck [y]$.
The relation $S$ is total (since $ \xck $ is total) and transitive
(since $ \xck $ is transitive): it is a total preorder.
It extends $R$ by the definition of $ \xec $
and the fact that $ \xck $ extends $ \xec$.
Let us show that it satisfies Equation (2) of Lemma 2.1.
Suppose $xSy$ and $ySx$.
We have $[x] \xck [y]$ and $[y] \xck [x]$
and therefore $[x]=[y]$ by antisymmetry
( $\xck$ is an order relation).
Therefore $x \xDd y$ and $xR^{*}y$.
$ \xcz $
\\[3ex]

The algebraic representation results we are going to demonstrate in this
Section 2 are independent of logic, and work for
arbitrary sets $U,$ not only for sets of models. On the other hand, if the
(propositional) language $ \xdl $ is defined from infinitely many
propositional
variables, not all sets of models are definable by a theory: There are $X
\xcc M_{ \xdl }$
s.t. there is no $T$ with $X=M(T).$ Moreover, we will consider only
consistent
theories. This motivates the following:

Let $U \xEd \xCQ,$ and let $ \xdy \xcc \xdp (U)$ contain all singletons,
be closed under finite non-empty $ \xcs $
and finite $ \xcv,$ $ \xCQ \xce \xdy $ and consider an operation $ \xfA:
\xdy \xCK \xdy \xcp \xdy.$ (For our
representation results, finite $ \xcs $ suffices.)

We are looking for a characterization of representable operators.
We first characterize those operations $ \xfA $ which can be represented
by a
symmetric pseudo-distance in Section 2.2, and then those representable by
a
not necessarily symmetric pseudo-distance in Section 2.3.

\bn

$\hspace{0.5em}$


For $a \xbe U,$ $X \xbe \xdy $ $a \xfA X$ will stand for $\{a\} \xfA X$
etc.

\en

\subsection{
The result for symmetric pseudo-distances
}


We work here with possibly infinite, but nonempty $U.$

Both Example 2.1 and Example 2.2 show that revision operators are
relatively
coarse instruments to investigate distances. The same revision operation
can be
based on many different distances. Consequently, in the construction of
the
distance from the revision operation, one still has a lot of freedom left.
Example 2.2 will show that, in the case one does not require
symmetric distances, the freedom is even greater.
The reader should
note that the situation described in Example 2.2 corresponds to the remark
in the proof of Proposition 2.5, that the constructed distance $d$ does
not
necessarily satisfy $d(A,B)=min\{d(a,B):a \xbe A\},$ i.e., may behave
strangely on the
left hand side.
But even when the pseudo-distance is a real distance, the
resulting revision operator $ \xfA_{d}$ does not always permit
reconstructing the
relations of the distances.

Distances with common start (or end, by symmetry) can always be
compared by looking at the result of revision:

$a \xfA_{d}\{b,b' \}=b$ iff $d(a,b)<d(a,b' ),$

$a \xfA_{d}\{b,b' \}=b' $ iff $d(a,b)>d(a,b' ),$

$a \xfA_{d}\{b,b' \}=\{b,b' \}$ iff $d(a,b)=d(a,b' ).$

This is not the case with arbitrary distances $d(x,y)$ and $d(a,b),$
as the following example will show.

\be

$\hspace{0.5em}$


We work in the real plane, with the standard distance, the angles have 120
degrees. $a' $ is closer to $y$ than $x$ is to $y,$ a is closer to $b$
than $x$ is to $y,$
but $a' $ is farther away from $b' $ than $x$ is from $y.$ Similarly
for b,b'.
But we cannot distinguish the situation $\{a,b,x,y\}$ and the
situation $\{a',b',x,y\}$ through $ \xfA_{d}.$
(See Figure 2.1.)

\ee

\paragraph{
Proof:
}

$\hspace{0.5em}$

Seen from $a$, the distances are in that order: $y,b,x$.

Seen from $a',$ the distances are in that order: $y,b',x$.

Seen from $b,$ the distances are in that order: $y,a,x$.

Seen from $b',$ the distances are in that order: $y,a',x$.

Seen from $y,$ the distances are in that order: $a/b,x$.

Seen from $y,$ the distances are in that order: $a'/b',x$.

Seen from $x,$ the distances are in that order: $y,a/b$.

Seen from $x,$ the distances are in that order: $y,a'/b'$.

Thus, any $c \xfA_{d}C$ will be the same in both situations (with $a$
interchanged with
$a',$ $b$ with $b'$). The same holds for any $X \xfA_{d}C$ where $X$ has
two elements.

Thus, any $C \xfA_{d}D$ will be the same in both situations, when we
interchange $a$ with
$a',$ and $b$ with $b'.$ So we cannot determine by $ \xfA_{d}$
whether $d(x,y)>d(a,b)$ or not.
$ \xcz $
\\[3ex]


\bfc
\begin{picture}(110,90)
\put(0,0){\usebox{\ZWEIeins}}
\end{picture}
\efc

\bp

$\hspace{0.5em}$


Let $U \xEd \xCQ,$ $ \xdy \xcc \xdp (U)$ be closed under finite
non-empty $ \xcs $
and finite $ \xcv,$ $ \xCQ \xce \xdy.$

Let $A,B,X_{i} \xbe \xdy.$

Let $ \xfA: \xdy \xCK \xdy \xcp \xdy,$ and consider the conditions

$( \xfA 1)$ $A \xfA B \xcc B$

$( \xfA 2)$ $A \xcs B \xEd \xCQ $ $ \xcp $ $A \xfA B=A \xcs B$

$( \xfA S1)$ (Loop):
$(X_{1} \xfA (X_{0} \xcv X_{2})) \xcs X_{0} \xEd \xCQ,$
$(X_{2} \xfA (X_{1} \xcv X_{3})) \xcs X_{1} \xEd \xCQ,$
$(X_{3} \xfA (X_{2} \xcv X_{4})) \xcs X_{2} \xEd \xCQ,$
 \Xl.
$(X_{k} \xfA (X_{k-1} \xcv X_{0})) \xcs X_{k-1} \xEd \xCQ $
imply
$(X_{0} \xfA (X_{k} \xcv X_{1})) \xcs X_{1} \xEd \xCQ.$

(a) $ \xfA $ is representable by a symmetric pseudo-distance $d:U \xCK U
\xcp Z$ iff $ \xfA $
satisfies $( \xfA 1)$ and $( \xfA S1).$

(b) $ \xfA $ is representable by an identity-respecting symmetric
pseudo-distance
$d:U \xCK U \xcp Z$ iff $ \xfA $ satisfies $( \xfA 1),$ $( \xfA 2),$ and
$( \xfA S1).$

\ep

Note that $( \xfA 1)$ corresponds to $(*2),$ $( \xfA 2)$ to $(*3),$ $(*0)$
will hold trivially,
$(*1)$ holds by definition of $ \xdy $ and $ \xfA,$ $(*4)$ will be a
consequence of
representation. $( \xfA S1)$ corresponds to:
$d(X_{1},X_{0}) \xck d(X_{1},X_{2}),$ $d(X_{2},X_{1}) \xck
d(X_{2},X_{3}),$ $d(X_{3},X_{2}) \xck d(X_{3},X_{4}) \xck $  \Xl  $ \xck
d(X_{k},X_{k-1}) \xck d(X_{k},X_{0})$
$ \xcp $ $d(X_{0},X_{1}) \xck d(X_{0},X_{k}),$ and, by symmetry,
$d(X_{0},X_{1}) \xck d(X_{1},X_{2}) \xck $  \Xl  $ \xck d(X_{0},X_{k})$ $
\xcp $
$d(X_{0},X_{1}) \xck d(X_{0},X_{k}),$ i.e. transitivity of $\leq$,
or to absence of loops involving $<$.

We first show the hard direction via a number of auxiliary definitions and
lemmas (up to Fact 2.4). We assume all A,B etc. to be in $ \xdy,$ and $(
\xfA 1),$ $( \xfA S1)$
to hold from now on.

We first define a precursor $ \xFO A,B \xFO $ to the pseudo-distance
between A and $B,$
and a relation $ \xck $ on these $ \xFO A,B \xFO ' s.$ We then prove some
elementary facts
about $ \xFO  \Xl  \xFO $ and $ \xck $ in Fact 2.3. We extend $ \xck $ to
a total preorder $S$ using
Lemma 2.1, the pseudo-distances will be S-equivalence classes of the $
\xFO A,B \xFO ' s.$
It remains to show that the revision operation $ \xfA_{d}$ defined by this
pseudo-distance is the same as the operation we started with, this is
shown in
Fact 2.4.

\bd

$\hspace{0.5em}$


Set $ \xFO A,B \xFO \xck \xFO A,B' \xFO $ iff $(A \xfA B \xcv B' )
\xcs B \xEd \xCQ,$

set $ \xFO A,B \xFO < \xFO A,B' \xFO $ iff $ \xFO A,B \xFO \xck \xFO
A,B' \xFO,$ but not $ \xFO A,B \xFO \xcg \xFO A,B' \xFO.$

\ed

$ \xFO A,B \xFO $ is to be read as the pseudo-distance between $A$ and $B$
or between $B$ and $A$.
Recall that the pseudo-distance will be symmetric, so $ \xFO  \Xl  \xFO $
operates on the
unordered pair $\{A,B\}.$ Note that $A \xfA B \xEd \xCQ,$ by definition
of the function $ \xfA.$

Let $ \xck^{*}$ be the transitive closure of $ \xck,$ we write also
$<^{*}$ if it involves $<.$
Write $ \xFO a,B \xFO $ for $ \xFO \{a\},B \xFO $ etc.

The loop condition reads in the $ \xFO $-notation as follows:
$ \xFO X_{0},X_{1} \xFO \xck \xFO X_{2},X_{1} \xFO \xck \xFO X_{2},X_{3}
\xFO \xck \xFO X_{4},X_{3} \xFO \xck  \Xl  \xck \xFO X_{k},X_{k-1} \xFO
\xck \xFO X_{k},X_{0} \xFO $ $ \xcp $
$ \xFO X_{0},X_{1} \xFO \xck \xFO X_{0},X_{k} \xFO $

\bfa

$\hspace{0.5em}$


(1) $ \xFO A,B \xFO \xcK \xFO A,B' \xFO $ iff $ \xFO A,B' \xFO < \xFO
A,B \xFO $

(2) $B' \xcc B$ $ \xcp $ $ \xFO A,B \xFO \xck \xFO A,B' \xFO $

(3) There are no cycles of the forms
$ \xFO A,B \xFO \xck \xFO A,B' \xFO \xck  \Xl  \xck \xFO A,B'' \xFO
\xck \xFO A,B \xFO $ or
$ \xFO A,B \xFO \xck \xFO A,B' \xFO \xck  \Xl  \xck \xFO A'',B \xFO
\xck \xFO A,B \xFO $
involving $<.$
(The difference between the two cycles is that the first contains possibly
only variations on one side, of the form $ \xFO A,B'' \xFO \xck \xFO
A,B \xFO \xck \xFO A,B' \xFO,$ the second
one possibly only alternating variations, of the form $ \xFO A'',B
\xFO \xck \xFO A,B \xFO \xck \xFO A,B' \xFO.)$

(4) $b \xbe A \xfA B$ $ \xcp $ $ \xFO A,b \xFO \xck \xFO A,B \xFO $

(5) $b \xce A \xfA B,$ $b \xbe B$ $ \xcp $ $ \xFO A,B \xFO < \xFO A,b \xFO
$

(6) $ \xFO A,b \xFO \xck^{*} \xFO A,B \xFO,$ $b \xbe B$ $ \xcp $ $b \xbe
A \xfA B$

(7) $b \xbe A \xfA B,$ $a_{b} \xbe b \xfA A,$ $a_{b} \xbe A' \xcc A$
implies
(a) $b \xbe A' \xfA B,$
(b) $A' \xfA B \xcc A \xfA B.$

(8) $b \xbe A \xfA B,$ $a_{b} \xbe b \xfA A,$ $a' \xbe A,$ $b' \xbe B$
$ \xcp $ $ \xFO a_{b},b \xFO \xck^{*} \xFO a',b' \xFO $

(9) $b \xbe B,$ $b \xce A \xfA B,$ $b' \xbe A \xfA B,$ $a_{b' } \xbe
b' \xfA A,$ $a \xbe A.$ Then $ \xFO a_{b' },b' \xFO <^{*} \xFO a,b
\xFO.$

If $( \xfA 2)$ holds, then

(10) $A \xcs B \xEd \xCQ $ $ \xcp $ $ \xFO A,B \xFO \xck^{*} \xFO A'
,B' \xFO $

(11) $A \xcs B \xEd \xCQ,$ $A' \xcs B' = \xCQ $ $ \xcp $ $ \xFO A,B
\xFO <^{*} \xFO A',B' \xFO $

\efa

\paragraph{
Proof:
}

$\hspace{0.5em}$

(1) and (2) are trivial.

(3) We prove both variants simultaneously.
Case 1, length of $cycle=1:$
$ \xFO A,B \xFO < \xFO A,B \xFO,$ so $(A \xfA B) \xcs B= \xCQ,$
contradiction.
Case 2: length $>1:$
Let e.g. $ \xFO A_{0},B_{0} \xFO \xck \xFO A_{0},B_{1} \xFO \xck  \Xl
\xck \xFO A_{0},B_{k} \xFO < \xFO A_{0},B_{0} \xFO $ be such a cycle.
If the cycle is not yet in the form of the loop condition, we can build
a loop as in the loop condition by repeating elements, if necessary. E.g.:
$ \xFO A_{0},B_{0} \xFO \xck \xFO A_{0},B_{1} \xFO \xck \xFO A_{0},B_{2}
\xFO $ can be transformed to
$ \xFO A_{0},B_{0} \xFO \xck \xFO A_{0},B_{1} \xFO \xck_{by \xDl (2)} \xFO
A_{0},B_{1} \xFO \xck \xFO A_{0},B_{2} \xFO.$ By Loop, we conclude
$ \xFO A_{0},B_{0} \xFO \xck \xFO A_{0},B_{k} \xFO,$ contradicting (1).

(4) and (5) are trivial.

(6) $b \xce A \xfA B$ $ \xcp_{by \xDl (5)} \xFO A,B \xFO < \xFO A,b \xFO
,$ contradicting $ \xFO A,b \xFO \xck^{*} \xFO A,B \xFO $ by (3).

(7) (a) By (6), it suffices to show that $ \xFO A',b \xFO \xck^{*} \xFO
A',B \xFO.$ But
$ \xFO A',b \xFO \xck_{by \xDl (2)} \xFO a_{b},b \xFO \xck^{*}_{(4)
\xDl twice} \xFO A,B \xFO \xck_{by \xDl (2)} \xFO A',B \xFO.$
(b) Let $b' \xbe A' \xfA B,$ we show $b' \xbe A \xfA B.$ By (6), it
suffices to show $ \xFO A,b' \xFO \xck^{*} \xFO A,B \xFO:$
$ \xFO A,b' \xFO \xck_{(2)} \xFO A',b' \xFO \xck_{(4)} \xFO A',B
\xFO \xck^{*}_{(2) \xDl twice} \xFO a_{b},b \xFO \xck^{*}_{(4) \xDl twice}
\xFO A,B \xFO.$

(8) $ \xFO a_{b},b \xFO \xck^{*} \xFO A,B \xFO \xck^{*} \xFO a',b'
\xFO.$

(9) $ \xFO a_{b' },b' \xFO \xck^{*}_{(4) \xDl twice} \xFO A,B \xFO
<_{(5)} \xFO A,b \xFO \xck_{(2)} \xFO a,b \xFO.$

(10) $ \xFO A,B \xFO \xck \xFO A,B \xcv B' \xFO,$ as $(A \xfA B \xcv
B' ) \xcs B \xEd \xCQ,$ by $A \xcs B \xcc A \xfA B \xcv B'.$
Likewise
$ \xFO A,B \xcv B' \xFO \xck \xFO A \xcv A',B \xcv B' \xFO.$
Moreover, $ \xFO A \xcv A',B \xcv B' \xFO \xck \xFO A',B' \xFO $
by (2).

(11) We show first that $A \xcs B \xEd \xCQ,$ $A \xcs B' = \xCQ $
implies $ \xFO A,B \xFO < \xFO A,B' \xFO:$
$A \xfA B \xcv B' =A \xcs (B \xcv B' )=A \xcs B \xcc A,$ so $(A \xfA B
\xcv B' ) \xcs B' = \xCQ.$
Thus, $ \xFO A,B \xFO \xck^{*}_{by \xDl (10)} \xFO A',A' \xFO < \xFO
A',B' \xFO.$
$ \xcz $
\\[3ex]

We define:

\bd

$\hspace{0.5em}$


Let $S,$ by Lemma 2.1, be a total preorder on $\{ \xFO A,B \xFO:A,B \xbe
\xdy \}$ extending $ \xck $ s.t.
$ \xFO A,B \xFO S \xFO A',B' \xFO $ and $ \xFO A',B' \xFO S \xFO
A,B \xFO $ imply $ \xFO A,B \xFO \xck^{*} \xFO A',B' \xFO.$

Let $ \xFO A,B \xFO \xCd \xFO A',B' \xFO $ iff $ \xFO A,B \xFO S \xFO
A',B' \xFO $ and $ \xFO A',B' \xFO S \xFO A,B \xFO,$ and $[
\xFO A,B \xFO ]$
be the set of $ \xCd $-equivalence classes and define $[ \xFO A,B \xFO ]<[
\xFO A',B' \xFO ]$ iff
$ \xFO A,B \xFO S \xFO A',B' \xFO $ but not $ \xFO A',B' \xFO S
\xFO A,B \xFO.$ This is a total order on
$\{[ \xFO A,B \xFO ]:$ $A,B \xbe \xdy \}.$ Define $d(A,B):=[ \xFO A,B \xFO
]$ for $A,B \xbe \xdy.$

If $( \xfA 2)$ holds, let $0:=[ \xFO A,A \xFO ]$ for any A. This is then
well-defined by
Fact 2.3, (10).

Note that by abuse of notation, we use $ \xck $ also between equivalence
classes.

\ed

\bfa

$\hspace{0.5em}$


(1) The restriction to singletons of $d$ as just defined is a symmetric
pseudo-distance; if $( \xfA 2)$ holds, then $d$ respects identity.

(2) $A \xfA B=A \xfA_{d}B.$

\efa

\paragraph{
Proof:
}

$\hspace{0.5em}$

(1)

(d1) Trivial.
If $[ \xFO b,c \xFO ]<[ \xFO a,a \xFO ],$ then $ \xFO b,c \xFO \xck^{*}
\xFO a,a \xFO,$ but not
$ \xFO a,a \xFO \xck^{*} \xFO b,c \xFO,$ contradicting Fact 2.3, (10).

(d2) $d(a,b)=d(a,a)$ iff $ \xFO a,b \xFO \xck^{*} \xFO a,a \xFO $ iff
$a=b$ by Fact 2.3, (10) and (11).

(d3) $[ \xFO a,b \xFO ] \xck [ \xFO b,a \xFO ]$ is trivial.

(2)

$`` \xcc '':$
Let $b \xbe A \xfA B.$ Then there is $a_{b} \xbe b \xfA A.$ By Fact 2.3,
(8), $ \xFO a_{b},b \xFO \xck^{*} \xFO a',b' \xFO $ for all
$a' \xbe A,$ $b' \xbe B.$ So $d(a_{b},b) \xck d(a',b' )$ for all
$a' \xbe A,$ $b' \xbe B$ and $b \xbe A \xfA_{d}B.$

$`` \xcd '':$
Let $b \xbe B,$ $b \xce A \xfA B.$ Take $b' \xbe A \xfA B,$ $a_{b' }
\xbe b' \xfA A,$ $a \xbe A.$ Then by Fact 2.3, (9)
$ \xFO a_{b' },b' \xFO <^{*} \xFO a,b \xFO,$ so $b \xce A \xfA_{d}B.$

$ \xcz $
\\[3ex]

It remains to show the easy direction of Proposition 2.2.

All conditions but $( \xfA S1)$ are trivial.
Define for two sets $A,B \xEd \xCQ $ $d(A,B):=d(a_{b},b),$ where $b \xbe A
\xfA_{d}B,$ and $a_{b} \xbe b \xfA_{d}A.$
Then $d(A,B)=d(B,A)$ by $d(a,b)=d(b,a)$ for all $a,b$. Loop amounts thus to
$d(X_{1},X_{0}) \xck  \Xl  \xck d(X_{k},X_{0})$ $ \xcp $ $d(X_{0},X_{1})
\xck d(X_{0},X_{k}),$ which is now obvious.
$ \xcz $ (Proposition 2.2)
\\[3ex]

\subsection{
The result for not necessarily symmetric pseudo-distances
}


Note that we work here with finite $U$ only, $ \xdy $ will be $ \xdp
(U)-\{ \xCQ \}.$

We first give an Example, which illustrates the expressive weakness of a
not
necessarily symmetric distance.

\be

$\hspace{0.5em}$


This example, illustrated in Figure 2.2, shows that we cannot find out, in
the
non symmetric case, which of the elements $a$, $a' $ is closest to the the
set $\{b,b' \}$
(we look from $a/a'$ to $\{b,b' \}).$ In the first case, it is
$a',$ in the second case $a$. Yet all results about revision stay the
same.

In the first case, we can take the ``road'' in both directions, in the
second case, we have to follow the arrows. (For simplicity, the vertical
parts
have length 0.) Otherwise, distances are as indicated by the numbers,
so e.g. in the second case, from $a' $ to $a$ it is 1, from $a$ to $a' $
1.2.
For any $X,Y \xcc \{a,a',b,b' \}$ $X \xfA Y$ will be the same in both
cases, but, seen from
$a$ or $a',$ the distance to $\{b,b' \}$ is closer from $a' $ in the
first case, closer from
$a$ in the second.

\ee


\bfc
\begin{picture}(110,80)
\put(0,0){\usebox{\ZWEIzwei}}
\end{picture}
\efc

The characterization of the not necessarily symmetric case presented in
the
following perhaps does not seem
very elegant at first sight, but it is straightforward and very useful
in the search for more elegant characterizations of similar operations.
For our characterization a definition is necessary.
It associates a binary relation between pairs of non-empty subsets of $U:$
intuitively, $(A,B)R_{ \xfA }(A',B' )$ may be understood as meaning
that the revision $\xfA$ requires the pseudo-distance between $A$ and $B$
to be smaller than or equal to that between $A' $ and $B'$.
The main idea of the representation theorem is to define
a relation (the relation $R_{ \xfA }$ of Definition 2.3)
that describes all inequalities we know must hold
between pseudo-distances, and require that the consequences of those
inequalities are upheld (conditions $( \xfA A2)$ and $( \xfA A3)$ of
Proposition 2.5).
The proof of the theorem shows that Definition 2.3 was comprehensive
enough.

\bd

$\hspace{0.5em}$


Given an operation $ \xfA,$ one defines a relation $R_{ \xfA }$
on pairs of non-empty subsets of $U$ by:
$(A,B)R_{ \xfA }(A',B' )$ iff one of the following two cases
obtains:

(1) $A=A' $ and $(A \xfA B \xcv B' ) \xcs B \xEd \xCQ,$

(2) $B=B' $ and $(A \xcv A' \xfA B) \xEd (A' \xfA B),$

If the pseudo-distance is to respect identity, we also consider a third
case:

(3) $A \xcs B \xEd \xCQ.$

\ed

Definition 2.3 can be written as:

(1) $(A \xfA B \xcv B' ) \xcs B \xEd \xCQ $ $ \xch $ $(A,B)R_{ \xfA
}(A,B' ),$

(2) $(A \xcv A' \xfA B) \xEd (A' \xfA B)$ $ \xch $ $(A,B)R_{ \xfA
}(A',B),$

(3) $A \xcs B \xEd \xCQ $ $ \xch $ $(A,B)R_{ \xfA }(A',B' ).$

In the sequel we shall write $R$ instead of $R_{ \xfA }.$
As usual, we shall denote by $R^{*}$ the reflexive and transitive closure
of $R.$

Notice also that we do not require that
the pseudo-distance between $A$ and $B$ be less or equal than that between
$A' $ and $B' $ if $A' \xcc A$ and $B' \xcc B,$ as one could
expect.
In fact, a theorem similar to Proposition 2.5 below may be proved
with a definition of $R$ that includes a fourth case:
$(A,B)R(A',B' )$ if $A' \xcc A$ and $B' \xcc B,$ and its proof is
slightly
easier, but we prefer to prove the stronger theorem.
Notice also that, in order to avoid the fourth case just mentioned,
the conclusion of case (2) is $(A,B)R_{ \xfA }(A',B),$
and not the seemingly stronger but in fact weaker in the absence
of the fourth case mentioned above: $(A,B)R_{ \xfA }(A \xcv A',B).$

We may now formulate our main technical result.
Condition $( \xfA A1)$ expresses a property of Disjunctive Rationality
([KLM90], [LM92], [Fre93]) for the left-hand-side argument
of the operation $ \xfA.$

\bp

$\hspace{0.5em}$


Consider the following conditions:

$( \xfA 1)$ $(A \xfA B) \xcc B,$

$( \xfA A1)$ $(A \xcv A' \xfA B) \xcc (A \xfA B) \xcv (A' \xfA B),$

$( \xfA A2)$ If $(A,B)R^{*}(A,B' ),$ then $(A \xfA B) \xcc (A \xfA B
\xcv B' ),$

$( \xfA A3)$ If $(A,B)R^{*}(A',B),$ then $(A \xfA B) \xcc (A \xcv A'
\xfA B),$

$( \xfA 2)$ If $A \xcs B \xEd \xCQ,$ then $A \xfA B=A \xcs B,$

$( \xfA A4)$ If $(A,B)R^{*}(A',B' )$ and $A' \xcs B' \xEd \xCQ,$
then $A \xcs B \xEd \xCQ.$

(a) An operation $ \xfA: \xdy \xCK \xdy \xcp \xdy $ is representable by a
pseudo-distance iff it satisfies the conditions $( \xfA 1),$ $( \xfA A1)-(
\xfA A3)$
for any non-empty sets $A,B \xcc U,$ where the relation $R$ is generated
by cases (1)
and (2) of Definition 2.3.

(b) An operation $ \xfA: \xdy \xCK \xdy \xcp \xdy $ is representable by
an identity-respecting
pseudo-distance iff it satisfies the conditions $( \xfA 1),$ $( \xfA 2),$
$( \xfA A1)-( \xfA A4)$
for any non-empty sets $A,B \xcc U,$ where the relation $R$ is generated
by cases (1) - (3)
of Definition 2.3.

\ep

\paragraph{
Proof:
}

$\hspace{0.5em}$

First, we shall deal with the soundness part of the theorem, and
then with the more challenging completeness part. We prove (a) and (b)
together.

Suppose, then, that $ \xfA $ is representable by a pseudo-distance.
The function $d$ acts on pairs of elements of $U,$ and it may be extended
to a function on pairs of non-empty subsets of $U$ in the usual way:
$d(A,B)$ $=$ $min\{d(a,b):a \xbe A,b \xbe B\}.$

Then Equation (1) in Definition 1.6, defining representability, may be
written as:

(3) $A \xfA B$ $=$ $\{b \xbe B \xfA d(A,b)=d(A,B)\}.$

We must now show that the conditions of Proposition 2.5 hold.

Condition $( \xfA 1)$ is obvious.

Condition $( \xfA A1)$ holds since $d(A \xcv A',B)$ $=$
$min\{d(A,B),d(A',B)\}.$

Considering Definition 1.6 and the different cases of
Definition 2.3, we shall see that $(A,B)R(A',B' )$ implies $d(A,B)
\xck d(A',B' ).$
Case 1 is obvious. Let us treat case (2). Clearly $d(A \xcv A',B)$ $=$
$min\{d(A',B),d(A,B)\}.$
We shall show that if $d(A',B)<d(A,B),$ then $A \xcv A' \xfA B=A'
\xfA B.$
Suppose $d(A',B)<d(A,B).$ Then, $d(A \xcv A',B)=d(A',B)<d(A,B).$
Therefore $A \xcv A' \xfA B=A' \xfA B.$
Case 2 has been taken care of. If $d$ respects identity, Case 3 is
obvious.
We conclude that $(A,B)R^{*}(A',B' )$ implies that $d(A,B) \xck
d(A',B' ).$
Condition $( \xfA A2)$ holds because $d(A,B) \xck d(A,B' )$ implies
$d(A,B \xcv B' )=d(A,B).$
Condition $( \xfA A3)$ holds because $d(A,B) \xck d(A',B)$ implies $d(A
\xcv A',B)=d(A,B).$

It remains to show that $( \xfA 2)$ and $( \xfA A4)$ follow from respect
of identity:

Condition $( \xfA 2)$ holds because $A \xfA B$ $=$ $\{b \xbe B:$
$d(A,b)=d(A,B)=0\}$ if $A \xcs B \xEd \xCQ.$
Condition $( \xfA A4)$ holds because $d(A,B) \xck d(A',B' )=0$
implies $d(A,B)=0.$

For the other direction, we work, unless stated otherwise, in the base
situation, i.e. where at least conditions $( \xfA 1),$ $( \xfA A1)-( \xfA
A3)$ hold, and the
relation $R$ is generated by at least cases (1) and (2) of Definition 2.3.

In our proof, a number of lemmas will be needed.
These lemmas will be presented when needed, and their proof
inserted in the midst of the proof of Proposition 2.5.
Again, we first extend the relation $R$ on pairs $(A,B)$ to a total preorder
$S$
using Lemma 2.1, and use the S-equivalence classes as pseudo-distances.
Recall that the $d$ thus defined will behave nicely on the right hand
side, but
not necessarily on the left hand side: $d(A,B)=min\{d(A,b):b \xbe B\}$
will hold, but
not necessarily $d(A,B)=min\{d(a,B):a \xbe A\}.$ Again, it remains to show
that
the revision operation $ \xfA_{d}$ defined by this pseudo-distance is the
same as the
operation we started with, this is done in the rest of the proof.

First, a simple result, analogous to the Or rule of [KLM90].

\bl

$\hspace{0.5em}$


For any sets $A,A',B,$ $(A \xfA B) \xcs (A' \xfA B) \xcc A \xcv A' \xfA
B.$

\el

\paragraph{
Proof:
}

$\hspace{0.5em}$

Without loss of generality we may assume that $A \xfA B \xEd A \xcv A'
\xfA B.$
Then $(A',B)R(A,B)$ by case (2) of Definition 2.3, and
$A' \xfA B \xcc A \xcv A' \xfA B$ by condition $( \xfA A3)$ of
Proposition 2.5.
$ \xcz $
\\[3ex]

We consider the set $ \xdy \xCK \xdy $ and the binary
relation $R$ on this set defined from $ \xfA $ by Definition 2.3.
By Lemma 2.1, $R$ may be extended to a total preorder $S$ satisfying:

(4) $xSy, ySx  \xch $ $xR^{*}y.$

Let $Z$ be the totally ordered set of equivalence classes of
$ \xdy \xCK \xdy $ defined by the total preorder $S.$
The function $d$ sends a pair of subsets $A$, $B$ to its equivalence class
under $S.$

We shall define $d(a,b)$ as $d(\{a\},\{b\}).$
Notice that we have first defined a pseudo-distance between subsets of
$U,$
and then a pseudo-distance between elements of $U.$
It is only the pseudo-distance between elements that is required by
the definition of representability.
The pseudo-distance between subsets just defined must be used with caution
because it does not satisfy the property: $d(A,B)$ $=$ $min\{d(a,b):a \xbe
A,b \xbe B\}.$
It satisfies half of it, as stated in Lemma 2.7 below.

Clearly, $(A,B)R(A',B' )$ implies $d(A,B) \xck d(A',B' ).$
Equation (4) also implies that if $d(A,B)=d(A',B' ),$ then
$(A,B)R^{*}(A',B' ).$

The following argument prepares respect of identity. Suppose that $ \xfA $
satisfies
$( \xfA 2)$ and $( \xfA A4)$ too, and that $R$ was defined including case
(3) of Definition 2.3.
Defining $0:=d(A,A)$ for any $A \xbe \xdy,$ we see that
(a) 0 is well-defined: By definition, $(A,A)R(B,B)$ for any $A,B \xbe \xdy
.$
(b) there is no $d(B,C)<0:$ By definition again, $(A,A)R(B,C).$
(c) $d(A,B)=0$ iff $A \xcs B \xEd \xCQ:$ $A \xcs B \xEd \xCQ $ implies
$(A,B)R(A,A),$ so $d(A,B)=0.$
$d(A,B)=0$ implies $(A,B)S(A,A)S(A,B),$ so $(A,B)R^{*}(A,A),$ so $A \xcs B
\xEd \xCQ $ by $( \xfA A4).$

The next lemma shows that our pseudo-distance $d$ behaves nicely as far as
its second argument is concerned.

\bl

$\hspace{0.5em}$


For any $A,B$ $d(A,B)=min\{d(A,b):b \xbe B\}$

and

(5) $A \xfA B=\{b \xbe B \xfA d(A,b)=d(A,B)\}.$

\el

\paragraph{
Proof:
}

$\hspace{0.5em}$

(Remember the elements of $ \xdy $ are non-empty.)
Suppose $b \xbe B.$ Since $(A \xfA B \xcv \{b\}) \xcs B \xEd \xCQ $ by
condition $( \xfA 1)$ of Proposition 2.5,
$(A,B)R(A,b)$ by case (1) of Definition 2.3, and therefore $d(A,B) \xck
min\{d(A,b):b \xbe B\}.$
If $b \xbe A \xfA B,$ then $(A \xfA B) \xcs \{b\} \xEd \xCQ $ and, by
Definition 2.3, case (1),
$(A,b)R(A,B)$ and therefore $d(A,b)=d(A,B).$
We have shown that the left hand side of Equation (5)
is a subset of the right hand side.
Since $A \xfA B$ is not empty there is a $b \xbe A \xfA B$ and, by the
previous remark,
$d(A,B)=d(A,b)$ and therefore we conclude that $d(A,B)=min\{d(A,b):b \xbe
B\}.$

To see that the right hand side of Equation (5) is a subset of the left
hand
side, notice that $d(A,B)=d(A,b)$ implies $(A,b)R^{*}(A,B)$ and therefore,
by condition $( \xfA A2)$ of Proposition 2.5, $A \xfA b \xcc A \xfA B$ and
$b \xbe A \xfA B.$
$ \xcz $
\\[3ex]

To conclude the proof of (a), we must show that Equation (1) of Definition
1.6
holds. Suppose, first,
that $b \xbe B,$ $a \xbe A$ and $d(a,b) \xck d(a',b' )$ for any $a'
\xbe A,$ $b' \xbe B.$ By Lemma 2.7,
$b \xbe a \xfA B$ and $d(a,B) \xck d(a',B),$ for any $a' \xbe A.$

We want to show now that $b \xbe A \xfA B.$ We will show that, for any
$a' \xbe A,$
$b \xbe \{a,a' \} \xfA B.$ One, then, concludes that $b \xbe A \xfA B$
by Lemma 2.6,
remembering that $U$ is finite. Since $b \xbe a \xfA B,$ we may, without
loss of
generality, assume that $a \xfA B \xEd \{a,a' \} \xfA B.$ By case (2) of
Definition 2.3,
$d(a',B) \xck d(a,B).$ But we already noticed that $d(a,B) \xck d(a'
,B).$
We can therefore conclude that $d(a,B)=d(a',B),$ so
$(a,B)R^{*}(a',B),$ $a \xfA B \xcc \{a,a' \} \xfA B$ and finally that
$b \xbe \{a,a' \} \xfA B.$
We have shown that the right hand side of Equation (1)
is a subset of the left hand side.

We proceed to show that the left hand side of Equation (1) is a subset of
its
right hand side.

Suppose that $b \xbe A \xfA B.$ By condition $( \xfA 1)$ of Proposition
2.5, $b \xbe B.$
We want to show that
there exists an $a \xbe A$ such that $d(a,b) \xck d(a',b' )$ for any
$a' \xbe A,$ $b' \xbe B.$
Since the set $U$ is finite, it is enough to prove that,
changing the order of the quantifiers:

(6) $ \xcA a' \xbe A,b' \xbe B, \xcE a \xbe A$ such that $d(a,b) \xck
d(a',b' ).$

Indeed, if Equation (6) holds, we get some $a \xbe A$
for every pair $a',b'$, and we may take the $a$
for which $d(a,b)$ is minimal: it satisfies the required condition.
Since $A= \xcV \{\{a',x\}:x \xbe A\}$ (the right-hand side is a finite
union) and
$b \xbe A \xfA B,$ by condition $( \xfA A1)$ of Proposition 2.5, there is
some $x \xbe A$
such that $b \xbe \{a',x\} \xfA B.$
We distinguish two cases.
First, if $b \xbe a' \xfA B,$ by Lemma 2.7, $d(a',b) \xck d(a'
,b' )$ and we may take $a=a'.$ Second,
suppose that $b \xce a' \xfA B.$ We notice that, since $b \xbe \{a'
,x\} \xfA B,$ condition $( \xfA A1)$ of
Proposition 2.5 implies that $b \xbe x \xfA B.$ But $b \xce a' \xfA B$
also implies that
$\{a',x\} \xfA B \xEd a' \xfA B.$ By Definition 2.3, case (2),
$(x,B)R(a',B)$ and $d(x,B) \xck d(a',B).$
But, by Lemma 2.7, we have $d(x,b) \xck d(x,B)$ (since $b \xbe x \xfA B)$
and $d(a',B) \xck d(a',b' ).$
We conclude that $d(x,b) \xck d(a',b' ),$ and we can take $a=x.$ This
concludes the proof
of (a).

It remains to show the rest of (b), respect of identity, i.e. that $A \xcs
B \xEd \xCQ $
implies $A \xfA B=A \xcs B,$ under the stronger prerequisites.
Let $A \xcs B \xEd \xCQ.$ Then for $b \xbe B$ $d(A,B)=0=d(A,b)$ iff $b
\xbe A.$ So by Equation (5)
$A \xfA B=A \xcs B.$
$ \xcz $
\\[3ex]

\section{
The logical representation results
}


\subsection{
Introduction
}


We turn to (propositional) logic.

\bd

$\hspace{0.5em}$


A pseudo-distance $d$ between models is called definability
preserving iff $ \xfA_{d}$ is.

$d$ is called consistency preserving iff $M(T) \xfA_{d}M(T' ) \xEd \xCQ
$ for consistent $T,T'.$

\ed

The role of definability preservation in the context of preferential
models is
discussed e.g. in [Sch92], [ALS98] discusses a similar problem in the
revision
of preferential databases, and its solution. This solution requires
much
more complicated conditions, for this reason, we have not adopted it here.

Note that $ \xcm T \xcr Th(M(T)),$ and $T=Th(M(T))$ if $T$ is deductively
closed.
Moreover, $X=M(Th(X))$ if there is some $T$ s.t. $X=M(T),$ so if the
operation $ \xfA $
is definability preserving, and $T*T' =Th(M(T) \xfA M(T' )),$ then
$M(T*T' )=M(T) \xfA M(T' ).$

\subsection{
The symmetric case
}


We consider the following conditions for a revision function $*$ defined
for
arbitrary consistent theories on both sides.

$(*0)$ If $ \xcm T \xcr S,$ $ \xcm T' \xcr S',$ then $T*T' =S*S'
,$

$(*1)$ $T*T' $ is a consistent, deductively closed theory,

$(*2)$ $T' \xcc T*T',$

$(*3)$ If $T \xcv T' $ is consistent, then $T*T' =Cn(T \xcv T' ),$

$(*S1)$ $Con(T_{0},T_{1}*(T_{0} \xco T_{2})),$ $Con(T_{1},T_{2}*(T_{1}
\xco T_{3})),$ $Con(T_{2},T_{3}*(T_{2} \xco T_{4}))$  \Xl
$Con(T_{k-1},T_{k}*(T_{k-1} \xco T_{0}))$ imply $Con(T_{1},T_{0}*(T_{k}
\xco T_{1})).$
\\[3ex]

Note: $(*4)$ of Definition 1.2 is for free, i.e. a consequence of $(*S1)$
and the
other conditions.
\\[3ex]

The following Example 3.1, very similar to Example 1.1,
shows that, in general, a revision operation defined
on models via a pseudo-distance by $T*T':=Th(M(T) \xfA_{d}M(T' ))$
will not satisfy $(*S1),$
unless we require $ \xfA_{d}$ to preserve definability.

\be

$\hspace{0.5em}$


Consider an infinite propositional language $ \xdl.$ We reinterpret the
models of
Example 1.1 as follows:

Let $T,$ $T_{1},$ $T_{2}$ be complete (consistent) theories, $T' $ a
theory with infinitely
many models, $T,T',T_{2}$ pairwise inconsistent. Let $M(T):=\{m\},$
$M(T_{1}):=\{m_{1}\},$
$M(T_{2}):=\{m_{2}\},$ $M(T' )=X \xcv \{m_{1}\}$ and $Th(X)=T'.$ (See
Figure 1.1.)

Then $M(T) \xfA M(T' )=X,$ but $T*T':=Th(X)=T'.$ We easily verify
$Con(T,T_{2}*(T \xco T)),$ $Con(T_{2},T*(T_{2} \xco T_{1})),$
$Con(T,T_{1}*(T \xco T)),$ $Con(T_{1},T*(T_{1} \xco T' )),$
$Con(T,T' *(T \xco T)),$ and conclude by Loop (i.e. $(*S1))$
$Con(T_{2},T*(T' \xco T_{2})),$ which
is wrong.
$ \xcz $
\\[3ex]

\ee

We finally have

\bp

$\hspace{0.5em}$


Let $ \xdl $ be a propositional language.

(a) A revision operation $*$ is representable by a symmetric consistency
and
definability preserving pseudo-distance iff $*$ satisfies $(*0)-(*2),$
$(*S1).$

(b) A revision operation $*$ is representable by a symmetric consistency
and
definability preserving, identity-respecting pseudo-distance iff $*$
satisfies $(*0)-(*3),$ $(*S1).$

\ep

\paragraph{
Proof:
}

$\hspace{0.5em}$

We prove (a) and (b) together.

For the first direction, let $ \xdy:=\{M(T):$ $T$ a consistent $ \xdl$
-theory\}, and define
$M(T) \xfA M(T' ):=M(T*T' ).$

By $(*0),$ this is well-defined, $ \xfA $ is obviously definability
preserving, and by
$(*1),$ $M(T) \xfA M(T' ) \xbe \xdy.$

We show the properties of Proposition 2.2.
$( \xfA 1)$ holds by $(*2),$ if $(*3)$ holds, so will $( \xfA 2).$ $( \xfA
S1)$ holds by $(*S1):$
E.g. $(M(T_{1}) \xfA (M(T_{0}) \xcv M(T_{2}))) \xcs M(T_{0}) \xEd \xCQ $
iff
$(M(T_{1}) \xfA M(T_{0} \xco T_{2})) \xcs M(T_{0}) \xEd \xCQ $ iff (by
definition of $ \xfA )$
$M(T_{1}*(T_{0} \xco T_{2})) \xcs M(T_{0}) \xEd \xCQ $ iff
$Con(T_{1}*(T_{0} \xco T_{2}),T_{0}).$
By Proposition 2.2, $ \xfA $ can be represented by an - if $( \xfA 2)$
holds, identity
respecting - symmetric pseudo-distance $d,$
so $M(T*T' )=M(T) \xfA M(T' )=M(T) \xfA_{d}M(T' ),$ and $Th(M(T*T'
))=Th(M(T) \xfA_{d}M(T' )).$
As $T*T' $ is deductively closed, $T*T' =Th(M(T*T' )).$

Conversely, define $T*T':=Th(M(T) \xfA_{d}M(T' )).$ We use
Proposition 2.2. $(*0)$ and $(*1)$
will trivially hold. By $( \xfA 1),$ $(*2)$ holds, if $( \xfA 2)$ holds,
so will $(*3).$
As above, we see that $(*S1)$ holds by $( \xfA S1),$ where now
$(M(T_{1}) \xfA_{d}M(T_{0} \xco T_{2})) \xcs M(T_{0}) \xEd \xCQ $ iff
$M(T_{1}*(T_{0} \xco T_{2})) \xcs M(T_{0}) \xEd \xCQ $
by definability preservation.
$ \xcz $
\\[3ex]

\subsection{
The not necessarily symmetric case
}

Recall that we work here with a language defined by finitely many
propositional variables.

For the not necessarily symmetric case, we consider the
following conditions for a revision function $*$ defined for
arbitrary consistent theories on both sides.

$(*0)$ If $ \xcm T \xcr S,$ $ \xcm T' \xcr S',$ then $T*T' =S*S'
,$

$(*1)$ $T*T' $ is a consistent, deductively closed theory,

$(*2)$ $T' \xcc T*T',$

$(*3)$ If $T \xcv T' $ is consistent, then $T*T' =Cn(T \xcv T' ),$

$(*A1)$ $(S \xco S' )*T$ $ \xcl $ $(S*T) \xco (S' *T),$

$(*A2)$ If $(S,T)R^{*}(S,T' ),$ then $S*T \xcl S*(T \xco T' ),$

$(*A3)$ If $(S,T)R^{*}(S',T),$ then $S*T \xcl (S \xco S' )*T,$

$(*A4)$ If $(S,T)R^{*}(S',T' )$ and $Con(S',T' ),$ then
$Con(S,T).$

Where the relation $R$ is defined by

(1) If $Con(S*(T \xco T' ),T),$ then $(S,T)R(S,T' ),$

(2) If $(S \xco S' )*T \xEd S' *T,$ then $(S,T)R(S',T),$

and, in the identity-respecting case, in addition by

(3) If $Con(S,T),$ then $(S,T)R(S',T' ).$

Note that by finiteness, any pseudo-distance is automatically definability
preserving. We have

\bp

$\hspace{0.5em}$


Let $ \xdl $ be a finite propositional language.

(a) A revision operation $*$ is representable by a consistency
preserving pseudo-distance iff $*$ satisfies $(*0)-(*2),$ $(*A1)-(*A3),$
where the relation $R$ is defined from the first two cases.

(b) A revision operation $*$ is representable by a consistency
preserving, identity-respecting pseudo-distance iff $*$ satisfies
$(*0)-(*3),$ $(*A1)-(*A4),$ where the relation $R$ is defined from all
three cases.

\ep

\paragraph{
Proof:
}

$\hspace{0.5em}$

We show (a) and (b) together.

We first note: If $T*T' =Th(M(T) \xfA M(T' )),$ then by definability
preservation
in the finite case $M(T*T' )=M(T) \xfA M(T' ),$ so
$(M(S) \xfA M(T) \xcv M(T' )) \xcs M(T) \xEd \xCQ $ $ \xcj $ $Con(S*(T
\xco T' ),T)$ and
$(M(S) \xcv M(S' )) \xfA M(T) \xEd M(S' ) \xfA M(T)$ $ \xcj $ $(S \xco
S' )*T \xEd S' *T.$
Thus, the relation $R$ defined in Definition 2.3 between sets of models,
and the
relation $R$ as just defined between theories correspond.

For the first direction, let $ \xdy:=\{M(T):$ $T$ a consistent $ \xdl$
-theory\}, and define
$M(T) \xfA M(T' ):=M(T*T' ).$

By $(*0),$ this is well-defined, and by $(*1),$ $M(T) \xfA M(T' ) \xbe
\xdy.$

We show the properties of Proposition 2.5.
$( \xfA 1)$ holds by $(*2).$
$( \xfA A1):$ We show $(M(S) \xcv M(S' )) \xfA M(T) \xcc (M(S) \xfA
M(T)) \xcv (M(S' ) \xfA M(T)).$
By $(*A1),$ $(S \xco S' )*T$ $ \xcl $ $(S*T) \xco (S' *T),$ so $(M(S)
\xcv M(S' )) \xfA M(T)$ $=$ $M(S \xco S' ) \xfA M(T)$ $=$
$M((S \xco S' )*T)$ $ \xcc $ $M(S*T) \xcv M(S' *T)$ $=$ $(M(S) \xfA
M(T)) \xcv (M(S' ) \xfA M(T)).$
For $( \xfA A2):$ Let $(M(S),M(T))R^{*}(M(S),M(T' )),$ so by the
correspondence between
the relation $R$ between sets of models, and the relation $R$ between
theories,
$(S,T)R^{*}(S,T' ),$ so by $(*A2)$ $S*T \xcl S*(T \xco T' ),$ so $M(S)
\xfA M(T)$ $ \xcc $ $M(S) \xfA (M(T) \xcv M(T' )).$
$( \xfA A3):$ Similar, using $(*A3).$ If $*$ satisfies $(*3)$ and $(*A4)$
and $R$ is also
generated by case (3), then $( \xfA 2)$ and $( \xfA A4)$ will hold by
similar arguments.

Thus, by Proposition 2.5, there is an (identity-respecting)
pseudo-distance $d$
representing $ \xfA,$ $M(T*T' )=M(T) \xfA_{d}M(T' )$ holds, so by
deductive closure of $T*T' $,
$T*T' =Th(M(T) \xfA_{d}M(T' )).$

Conversely, define $T*T':=Th(M(T) \xfA_{d}M(T' )).$ We use again
Proposition 2.5. $(*0)$ and
$(*1)$ will trivially hold. The proof of the other properties closely
follows
the proof in the first direction.
$ \xcz $
\\[3ex]

\section{Conclusion}

We proposed a pseudo-distance semantics for the AGM theory of revision.
Our semantics is in line with AGM's identification of epistemic states
with belief sets.
It validates additional postulates that ensure coherence conditions
concerning the dependence of the revision operator $*$ on its
left argument.
Those postulates have been exactly characterized by representation results
in the case of symmetric pseudo-distances and only in the finite case for
general pseudo-distances.

The question of a representation theorem for the infinite case of general
pseudo-distances with conditions in our style stays open.
As our main results are purely
algebraic in
nature, one can be quite confident that important parts of our
constructions
can be used in the richer situation, in which epistemic states contain
more than belief sets.

\section{Acknowledgement}

We would like to thank a referee for his very insightful comments, which
have helped to improve the paper.

\end{document}